\documentclass[conference,10pt,twocolumn]{IEEEtran}

\usepackage{geometry}
\usepackage{graphicx}
\usepackage{amsmath,amssymb,amsthm}
\usepackage{algorithm}
\usepackage{algpseudocode}
\usepackage{booktabs}
\usepackage{multirow}
\usepackage{subcaption}
\usepackage{tikz}
\usetikzlibrary{shapes,arrows,positioning,calc,fit,backgrounds,decorations.pathreplacing,patterns,shadows,shapes.geometric,arrows.meta,matrix,decorations.markings,fadings}
\usepackage{pgfplots}
\pgfplotsset{compat=1.17}
\usepgfplotslibrary{fillbetween}

\definecolor{navyColor}{RGB}{27, 58, 92}
\definecolor{dataColor}{RGB}{65, 105, 145}
\definecolor{policyColor}{RGB}{70, 110, 100}
\definecolor{fusionColor}{RGB}{90, 75, 115}
\definecolor{outputColor}{RGB}{140, 120, 80}
\definecolor{successColor}{RGB}{70, 110, 80}
\definecolor{gapColor}{RGB}{140, 70, 70}
\definecolor{contextColor}{RGB}{100, 100, 100}
\definecolor{stage1Color}{RGB}{75, 95, 120}
\definecolor{stage2Color}{RGB}{90, 75, 115}
\definecolor{stage3Color}{RGB}{130, 95, 70}

\tikzset{
    stagebox/.style={rectangle, rounded corners=5pt, draw=#1!60, fill=#1!5, line width=0.8pt},
    eqbox/.style={rectangle, draw=#1!70, fill=white, rounded corners=2pt, align=center, inner sep=4pt, font=\tiny, line width=0.5pt},
    procbox/.style={rectangle, draw=#1, fill=#1!8, rounded corners=2pt, minimum width=2cm, minimum height=0.5cm, font=\scriptsize, line width=0.4pt, align=center},
    stagenum/.style={circle, draw=#1, fill=#1!15, minimum size=0.5cm, font=\scriptsize\bfseries, text=#1, line width=0.5pt},
    flowarrow/.style={->, >=stealth, line width=0.6pt, color=contextColor},
    stagearrow/.style={->, >=stealth, line width=1.2pt},
    factorbox/.style={rectangle, rounded corners=3pt, draw=#1, fill=#1!12, minimum width=2.1cm, minimum height=0.9cm, font=\tiny, align=center, line width=0.5pt},
}

\usepackage{hyperref}
\usepackage{xcolor}

\makeatletter
\g@addto@macro\UrlBreaks{\do\-\do\/\do\.\do\_\do\?\do\&\do\=\do\:}
\makeatother
\let\IEEEoldthebibliography\thebibliography
\let\IEEEoldendthebibliography\endthebibliography
\renewenvironment{thebibliography}[1]{%
    \IEEEoldthebibliography{#1}%
    \normalsize
}{%
    \IEEEoldendthebibliography
}

\usepackage{enumitem}
\usepackage{float}
\usepackage{array}
\usepackage{tabularx}

\newtheorem{definition}{Definition}

\newcolumntype{C}[1]{>{\raggedright\arraybackslash}p{#1}}
\newcolumntype{Y}{>{\raggedright\arraybackslash}X}

\title{Artificial Intelligence for Energy Optimization in Data Centers:\\
Closing the Optimizer--Load Loop}

\author{
\IEEEauthorblockN{Mohammed Basharath Ullah}
\IEEEauthorblockA{Hyderabad, India \\
mohdbasharath8@gmail.com}
\and
\IEEEauthorblockN{Summaiya Unnisa Begum}
\IEEEauthorblockA{Hyderabad, India \\
summaiyaunisa@gmail.com}
\and
\IEEEauthorblockN{Mohammed Nadeem Ullah}
\IEEEauthorblockA{Hyderabad, India \\
mdnadeemullah@gmail.com}}

\begin{document}

\maketitle

\section*{\textbf{Abstract}}

Data centers are increasingly optimized by artificial intelligence and, at the same time, increasingly loaded by it. The literature treats these as two unrelated problems: control studies model workload as an exogenous arrival process, while sustainability studies model infrastructure as a fixed multiplier. We screen roughly 194 papers retrieved through a documented protocol, code 63 of them, and report what the coding shows. Of 28 primary control-oriented studies, 18 are validated in simulation alone and 5 reach physical hardware or a production facility; none account for water withdrawal, and none account for embodied carbon. Reported savings intervals across four technique families overlap almost completely, which means the field cannot presently rank its own methods. Ten recurring gaps are scored for consequence and tractability, and we set out CLEAR-DC, a framework coupling a control-policy branch to a workload-demand branch through an explicit elasticity term, reads out net rather than direct benefit, and emits a schema-conformant record covering energy, carbon, water, embodied share and validation venue. The framework is an architectural and methodological proposal, not a trained system; the contribution we defend empirically is the corpus analysis and the reporting schema derived from it. Coding sheet, derived statistics and all result artifacts: {\footnotesize\url{https://github.com/Kimalice/AI-for-Energy-Optimization-in-Data-Centers-Closing-the-Optimizer-Load-Loop}}

\noindent\textbf{Keywords:} Data Center Energy Efficiency, Deep Reinforcement Learning, Cooling Optimization, Workload Scheduling, Carbon-Aware Computing, Sustainable AI, Rebound Effects, Systematic Review, Reporting Standards

\section{Introduction: Why Efficiency Claims Need a Closed Loop}

Digital infrastructure has become an energy sector in its own right. Early accounting placed data center electricity at roughly one percent of world consumption once cooling and power distribution were included \cite{ref20}, and subsequent bottom-up modelling showed that demand for services grew far faster than electricity use because efficiency improvements absorbed most of the difference \cite{ref50,ref31}. That reassurance is now under pressure. Forecasts diverge sharply depending on assumptions about hardware trends and service growth \cite{ref21}, with recent scenario work spanning roughly 1{,}800 to 5{,}000 TWh by mid-century \cite{ref10}, while a competing long-run analysis argues that computing's share of world electricity peaked over a decade ago and has since stabilised \cite{ref40}. The disagreement is not merely academic: it determines whether efficiency research is a marginal optimisation or a decarbonisation priority.

Against that backdrop, artificial intelligence entered the field first as a modelling tool and then as a controller. Neural network models learned facility behaviour directly from operational sensor data and predicted power usage effectiveness with error small enough to guide set-point selection \cite{ref13}. Deep reinforcement learning then closed the loop, replacing hand-tuned control logic with learned policies \cite{ref26}, and the control surface widened to span information technology and cooling subsystems jointly \cite{ref42,ref43}, to enforce thermal safety during learning \cite{ref55,ref4}, and to follow carbon intensity across time and geography \cite{ref41,ref46}. Reported savings are substantial and consistent enough that several reviews now treat the approach as established \cite{ref18,ref25,ref39}.

\subsection{A Commissioning Scenario}

Consider an operator commissioning a learned controller in a facility that also hosts large-scale model training. The controller is expected to reduce facility energy while the workload it manages is growing at a rate no cooling optimisation can offset, drawing water in a basin that may be stressed, and imposing power transients the facility's own models do not represent. Three obstacles frustrate any attempt to evaluate whether that deployment is a net improvement:

\begin{enumerate}[leftmargin=*]
    \item \textbf{Open-Loop Framing}: Control studies treat workload as an exogenous stochastic process and report the energy saved at fixed demand. Sustainability studies treat infrastructure efficiency as a fixed coefficient applied to a demand trajectory determined elsewhere. Neither models the path by which efficiency reduces the effective unit cost of computation and thereby influences how much computation is purchased. A systematic review spanning the artificial intelligence lifecycle arrives at a compatible position from a different direction, reporting that gains realised at one stage are repeatedly offset by responses elsewhere in the system, so that stage-level improvements do not aggregate into lifecycle ones \cite{ref38}. That offset is named there but not formalised, and it is the path it leaves open that this paper tries to close.
    \item \textbf{Validation Deficit}: Learned controllers are overwhelmingly evaluated in simulation. A review of reinforcement learning for building climate control found that under a quarter of studies reached a real building \cite{ref2}, and a systematic review focused on data centers names the absence of real-time validation as a principal gap \cite{ref18}. A dedicated evaluation across algorithms, tasks, system dynamics and transfer settings reports that leading methods remain sensitive to hyperparameters, reward specification and operating scenario, with constraint violations and sample efficiency still unsatisfactory \cite{ref62}.
    \item \textbf{Measurement Incomparability}: Savings are reported against heterogeneous baselines and boundaries. One review records energy reductions ranging from roughly 5 to 90 percent for heuristics, 8 to 97 percent for metaheuristics and 2 to 89 percent for machine learning \cite{ref39}, intervals so wide and so overlapping that no ranking survives them. The underlying evidence base is itself uneven: an audit of 258 published estimates found under a third of their sources peer-reviewed \cite{ref35}, and global figures for a single year have been reported across a sixfold range \cite{ref57}.
\end{enumerate}

\textbf{Research Question}: Can data center energy optimisation be evaluated in a way that accounts simultaneously for the direct savings a controller achieves, the induced demand its efficiency may create, the resources beyond electricity that it consumes, and the conditions under which its reported result can be compared with anyone else's? This paper addresses that question through a coded corpus analysis, a prioritised gap synthesis, and a framework designed against the gaps identified.

\begin{figure}[!t]
\centering
\resizebox{\columnwidth}{!}{%
\begin{tikzpicture}
\tikzset{
  driver/.style={rectangle, rounded corners=3pt, draw=#1, fill=#1!12,
                 text width=2.35cm, minimum height=1.15cm, font=\tiny,
                 align=center, line width=0.5pt, inner sep=2pt},
}
\node[driver=stage1Color] (it)    at (0.00, 3.20) {\textbf{IT workload}\\compute, GPU};
\node[driver=dataColor]   (cool)  at (2.70, 3.20) {\textbf{Cooling plant}\\CRAC, liquid};
\node[driver=navyColor]   (power) at (5.40, 3.20) {\textbf{Power supply}\\grid carbon mix};
\node[driver=policyColor] (water) at (0.00, 1.75) {\textbf{Water use}\\cooling + power};
\node[driver=outputColor] (emb)   at (2.70, 1.75) {\textbf{Embodied carbon}\\hardware, build};
\node[driver=stage3Color] (grid)  at (5.40, 1.75) {\textbf{Grid coupling}\\ramp, flexibility};

\draw[contextColor!60, line width=0.5pt] (-1.15,0.85) -- (6.55,0.85);
\foreach \n in {it,cool,power}{\draw[contextColor!50, line width=0.4pt] (\n.south) -- ($(\n.south)+(0,-0.30)$);}
\foreach \n in {water,emb,grid}{\draw[contextColor!50, line width=0.4pt] (\n.south) -- (\n.south |- 0,0.85);}

\node[rectangle, rounded corners=3pt, draw=gapColor, fill=gapColor!12,
      minimum width=5.0cm, minimum height=0.95cm, font=\small\bfseries,
      align=center, line width=0.7pt] (fp) at (2.70,-0.25)
      {Data centre energy\\and carbon footprint};
\draw[flowarrow] (2.70,0.85) -- (fp.north);

\node[rectangle, rounded corners=3pt, draw=stage3Color, fill=stage3Color!15,
      minimum width=5.0cm, minimum height=0.8cm, font=\small\bfseries,
      line width=0.7pt] (ai) at (2.70,-2.15) {AI optimisation layer};
\draw[flowarrow, color=stage3Color] (1.55,-0.73) -- (1.55,-1.75)
      node[midway, left, font=\tiny] {controls};
\draw[flowarrow, color=stage3Color] (3.85,-1.75) -- (3.85,-0.73)
      node[midway, right, font=\tiny] {measures};

\draw[->, >=stealth, line width=0.7pt, color=gapColor]
      (ai.west) -- (-1.50,-2.15) -- (-1.50,3.20) -- (it.west);
\node[font=\tiny, text=gapColor, rotate=90] at (-1.85,0.6) {induced demand (unmodelled)};
\end{tikzpicture}}
\caption{Drivers of the data center footprint. The optimisation layer controls the facility, but the demand it makes cheaper feeds back into workload --- a path left unmodelled in the studies surveyed here.}
\label{fig:drivers}
\end{figure}
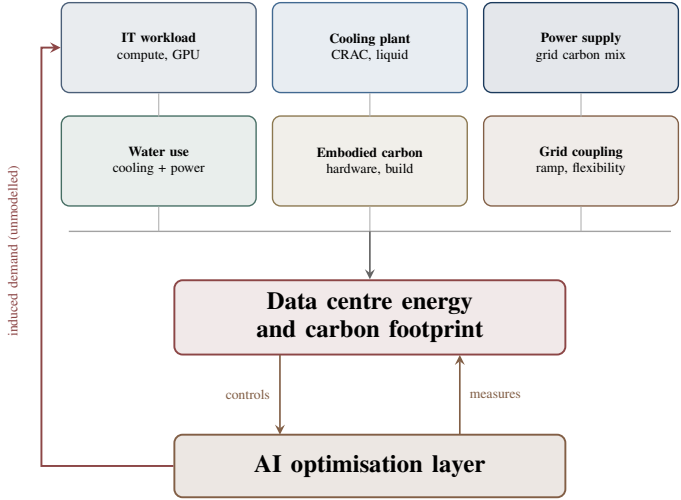

Figure~\ref{fig:drivers} summarises the drivers that jointly determine a facility's footprint and marks the feedback path that current formulations omit.

\subsection{What Is Missing, and What We Add}

Existing reviews of this field catalogue architectures and compare reported accuracy, but rarely code their corpus, and none we located quantify how the evidence base is distributed across validation venues or resource scopes. The contributions are listed below, each with the limits on what it claims:

\begin{itemize}[leftmargin=*]
    \item \textbf{Seven-Paradigm Synthesis with Gap Identification}: A comparative reading of measurement, prediction, cooling control, scheduling, carbon-aware computing, sustainable AI and enabling technologies (Section~II), each closed by naming what it leaves undone.
    \item \textbf{Coded Corpus Analysis}: Sixty-three papers coded by validation venue, scope, and resource breadth, with the resulting distribution reported in Section~VI. \textbf{Scope}: coding is performed at abstract level, not full text, and is single-coded rather than double-coded (see Section~VII).
    \item \textbf{Ranked Gap Set}: Ten deficiencies derived from the coding and ranked on two axes, how much each matters and how readily it could be closed (Section~III). \textbf{Scope}: both scores reflect our own reading of the corpus; no panel was convened to set them.
    \item \textbf{CLEAR-DC}: A framework coupling a control-policy branch to a workload-demand branch through an explicit elasticity term, traced component by component to the gaps it targets (Section~III). \textbf{Scope}: CLEAR-DC is specified here, not implemented, trained or measured.
    \item \textbf{Minimum Reporting Schema}: A set of fields that, reported together, would let future results be compared with one another; Section~IV sets each beside its present status.
\end{itemize}

The paper continues as follows. Section~II reviews the seven paradigms and their deficiencies. Section~III states the review protocol, formalises the problem and presents CLEAR-DC. Section~IV gives the evaluation design and an India-focused case. Section~V describes reproducibility. Section~VI reports the corpus analysis. Section~VII discusses limitations and impact, and Section~VIII concludes.

\section{Seven Paradigms and What Each Leaves Undone}

The corpus divides into seven paradigms. Each subsection below ends by naming the deficiency that the paradigm leaves behind, and it is those deficiencies that Section~III scores.

\subsection{Measurement and Macro-Estimation}
The oldest strand asks how much electricity data centers actually consume. Bottom-up accounting established the early baseline \cite{ref20} and was later refined to show that United States demand had flattened even as workloads grew \cite{ref50}, a finding generalised globally to argue that efficiency trends had offset service growth \cite{ref31}. Forecasting work extends the horizon under explicit socio-economic assumptions \cite{ref21,ref10}, while critical audits question the provenance of the numbers everyone cites \cite{ref35,ref57} and tutorial surveys have begun to gather infrastructure and sustainability concerns into one account \cite{ref8}. A parallel line quantifies footprints beyond electricity, including water withdrawal and its spatial concentration in stressed watersheds \cite{ref34,ref51,ref23,ref19}.

\textbf{Gap Identification}: This strand supplies the denominator but not the controller. Its resource breadth --- water, spatial stress, embodied impact --- has not propagated into the optimisation literature that cites it for motivation.

\subsection{Predictive Modelling}
Data-driven models of facility and workload behaviour preceded learned control. Neural models trained on operational data predicted power usage effectiveness accurately enough to guide configuration \cite{ref13}, and comparable methods were later deployed at hyperscale to tune cooling set points and detect unreasonable operating conditions \cite{ref60}. Workload forecasting matured in parallel, with systematic comparison of classical and deep predictors on common cluster traces \cite{ref48}.

\textbf{Gap Identification}: Prediction accuracy is rarely connected to realised energy outcomes. A model that forecasts utilisation well is reported as a forecasting result, leaving the downstream saving it enables unquantified.

\subsection{Deep Reinforcement Learning for Cooling}
Learned control began with off-policy actor-critic methods applied end to end to cooling plant, reporting cooling cost reductions in the low double digits against manually configured baselines \cite{ref26}. Joint optimisation of scheduling and airflow followed, using a parameterised action space to handle the mixed discrete-continuous decision and a two-timescale mechanism to reconcile subsystems that respond at different rates \cite{ref42,ref43}, later extended across geographically distributed sites \cite{ref44}. Multi-agent decomposition addressed the dimensionality of the joint action space \cite{ref6}. Safety then became a first-class constraint: imitation pretraining combined with post-hoc action rectification reduced thermal violations by a wide margin relative to reward shaping while still saving substantial facility power \cite{ref55}, and a Lyapunov-based formulation replaced the uniform-temperature assumption with a rack-level compliance metric \cite{ref4}. Surrogate modelling made real-time optimisation tractable by displacing computational fluid dynamics from the control loop \cite{ref63,ref29}.

\textbf{Gap Identification}: Deployment evidence has not kept pace with algorithmic sophistication. Sensitivity to reward design, hyperparameters and scenario shift is documented but unresolved \cite{ref62}, degradation under non-stationary conditions has been observed in the field \cite{ref56}, and simplified occupancy or load assumptions are known to inflate simulated performance \cite{ref33}.

\subsection{Workload Scheduling and Consolidation}
On the compute side, consolidation reduces server power by packing load onto fewer active machines. Prediction-aware policies that anticipate future utilisation rather than reacting to current utilisation reduce needless migrations and service-level violations \cite{ref11,ref15}, and deep reinforcement learning has largely displaced heuristics for placement \cite{ref61,ref14}. The central tension is thermal: aggressive packing concentrates heat and creates hotspots that erode the saving \cite{ref16}, a concern extended to heat recirculation in recent scheduling work \cite{ref24}. Server-level control combining frequency scaling with fan management is one of the few strands validated on physical hardware \cite{ref27}.

\textbf{Gap Identification}: Reported savings in this paradigm span nearly the entire feasible range across studies \cite{ref39}, and almost all evaluation is trace-driven simulation on a small number of public cluster traces, which makes the spread impossible to attribute to method rather than setting.

\subsection{Carbon-Aware and Grid-Interactive Computing}
A production system at hyperscale demonstrated that temporally flexible workloads can be delayed toward lower-carbon hours using day-ahead intensity forecasts and hourly capacity limits \cite{ref41}. Subsequent work extends flexibility across space as well as time \cite{ref46,ref59}, the latter explicitly pricing the carbon cost of migration rather than counting only its benefit. Data centers have since been reframed as flexibility assets for the grid \cite{ref52}, with empirical trace analysis quantifying how much load is genuinely deferrable \cite{ref5}, and robust or uncertainty-aware formulations addressing renewable variability \cite{ref1}.

\textbf{Gap Identification}: A review of this paradigm finds that spatial and temporal shifting are usually studied in isolation, with only a small minority combining them in one optimisation \cite{ref3}. More troubling, flexibility is not unambiguously beneficial: under low renewable penetration it can reduce system cost while increasing emissions \cite{ref49}.

\subsection{Sustainable AI: When AI Becomes the Load}
A distinct literature measures the footprint of the models themselves. Lifecycle accounting for a large language model separated operational training emissions from the substantially larger total once manufacturing and surrounding processes were included \cite{ref28}, a distinction formalised into a projection tool covering dense and sparse architectures \cite{ref9}. A systematic review mapped the field and found it concentrated on the training phase \cite{ref53} --- a focus since overtaken by evidence that inference efficiency depends heavily on workload geometry, software stack and accelerator, and that theoretical operation counts materially understate real consumption \cite{ref12}. Holistic evaluation showed that undisclosed model development contributed a large share of total impact alongside significant water use \cite{ref32}, and architectural analysis argued for optimising across the whole infrastructure lifecycle rather than efficiency alone \cite{ref58}. Direct measurement of accelerator nodes during training provides the hardware-level ground truth \cite{ref22}, while power stabilisation work shows that synchronised training produces transients whose frequency content can threaten grid equipment \cite{ref7}. A dissenting analysis argues that per unit of output, model inference can be far less impactful than the human labour it substitutes \cite{ref45}.

\textbf{Gap Identification}: This paradigm and the control paradigms above are effectively disjoint. Neither cites the other's mechanisms, and no reviewed study lets an efficiency result in one feed back into a demand assumption in the other.

\subsection{Enablers: Digital Twins, Benchmarks and Explainability}
Three enabling technologies determine whether the preceding work becomes deployable. Automated calibration of facility models against sensor measurements removed a manual bottleneck and achieved sub-degree agreement on production halls \cite{ref54}. Shared benchmark environments appeared only recently, first covering scheduling, cooling and battery management jointly \cite{ref36}, then extending to liquid cooling with interpretable and language-model-based control baselines \cite{ref37} and to hierarchical control across facility clusters \cite{ref47}. Explainability, by contrast, is mature in adjacent power-system research \cite{ref30} but has barely transferred, even though reviews of data center optimisation list transparent decision-making among their unresolved challenges \cite{ref25}.

\textbf{Gap Identification}: Because shared benchmarks arrived in 2024 and 2025, effectively the entire preceding decade was evaluated on private, non-reproducible configurations, and no retrospective re-evaluation of headline results on common environments has been published.

\subsection{Where This Work Sits}

Table~\ref{tab:related_work_comparison} sets the proposed direction beside representative prior work on five axes: joint optimisation of compute and cooling (Joint), explicit safety guarantees during learning (Safe), resource breadth beyond electricity and carbon (Res), closure of the optimiser--load feedback loop (Loop), and validation beyond simulation (Real).

\begin{table}[!t]
\centering
\caption{Coverage of representative prior work on five dimensions.}
\label{tab:related_work_comparison}
\small
\setlength{\tabcolsep}{3pt}
\renewcommand{\arraystretch}{1.15}
\begin{tabular}{@{}>{\raggedright\arraybackslash}p{2.6cm}ccccc@{}}
\toprule
\textbf{Paradigm} & \textbf{Joint} & \textbf{Safe} & \textbf{Res} & \textbf{Loop} & \textbf{Real} \\
\midrule
Measurement \cite{ref31,ref35} & & & $\circ$ & & $\checkmark$ \\
Predictive ML \cite{ref13,ref48} & & & & & $\checkmark$ \\
DRL cooling \cite{ref26,ref43} & $\checkmark$ & & & & \\
Safe RL \cite{ref55,ref4} & $\checkmark$ & $\checkmark$ & & & \\
Consolidation \cite{ref11,ref16} & $\circ$ & & & & \\
Carbon-aware \cite{ref41,ref46} & $\circ$ & & & & $\checkmark$ \\
Sustainable AI \cite{ref28,ref58} & & & $\checkmark$ & & $\checkmark$ \\
Benchmarks \cite{ref36,ref37} & $\checkmark$ & $\circ$ & $\circ$ & & \\
\midrule
\textbf{Ours (CLEAR-DC)} & $\checkmark$ & $\checkmark$ & $\checkmark$ & $\checkmark$ & $\circ$ \\
\bottomrule
\end{tabular}

\vspace{2pt}
{\small $\checkmark$ addressed; $\circ$ partially addressed. The Real
column for CLEAR-DC is marked partial because the framework specifies a
validation-venue field rather than itself reporting a deployment.}
\end{table}

\section{Method: Protocol, Gap Scoring and Framework Design}

\subsection{Review Protocol}

Retrieval was performed over an index aggregating Semantic Scholar, PubMed, Scopus and arXiv. Twenty queries were executed, structured as one exploratory query, five sub-area queries, three secondary-study queries restricted to review and systematic-review types, two era-gated queries bounding publication year, and nine targeted queries covering water, grid interaction, accelerator power, workload forecasting, explainability and benchmark environments. Each query returned at most ten records, giving 199 records and approximately 194 unique papers after deduplication by title.

Screening retained records that addressed data center or cloud energy consumption, its optimisation, or the footprint of the workloads themselves, and that reported either a method, a measurement or a synthesis. Sixty-three papers were retained and coded. Each was assigned one validation venue (simulation, analytical model, measurement study, physical testbed or field deployment, production facility, or secondary study), one primary scope (cooling, scheduling, carbon-aware, AI-as-load, enabler, or macro-estimation), and two binary flags recording whether water withdrawal and embodied carbon were accounted for. Where a paper spanned scopes, the scope corresponding to its principal experimental contribution was recorded.

\subsection{Notation and Symbols}

Symbols recur throughout Sections~III to~VI; Table~\ref{tab:notation} defines each one once.

\begin{table}[!t]
\centering
\caption{Notation for Sections~III--VI.}
\label{tab:notation}
\small
\setlength{\tabcolsep}{4pt}
\renewcommand{\arraystretch}{1.15}
\begin{tabular}{@{}>{$}l<{$}p{6.20cm}@{}}
\toprule
\textnormal{\textbf{Symbol}} & \textbf{Description} \\
\midrule
\pi, \pi_0 & Learned control policy and incumbent baseline controller \\
\mu_0 & Workload demand process under the baseline \\
\mu_\varepsilon & Workload demand after efficiency-induced adjustment \\
\varepsilon & Elasticity of compute demand to effective unit cost \\
\delta & Relative reduction in effective unit cost of compute \\
E(\pi, \mu) & Facility energy under policy $\pi$ and demand $\mu$ \\
C, W & Operational carbon and water withdrawal, same arguments \\
\Xi & Embodied carbon attributed to the evaluation window \\
\Phi & Thermal-safety constraint set (e.g. rack cooling index) \\
\Delta E_{\mathrm{net}} & Net energy benefit after induced demand \\
\sigma & Reporting record emitted for one evaluation \\
\nu & Validation venue label attached to $\sigma$ \\
\bottomrule
\end{tabular}
\end{table}

\subsection{Scoring the Gaps}

\begin{figure}[!t]
\centering
\begin{tikzpicture}
\begin{axis}[
    width=8.4cm, height=6.2cm,
    xlabel={Feasibility (1--5)}, ylabel={Impact (1--5)},
    xlabel style={font=\scriptsize}, ylabel style={font=\scriptsize},
    xmin=2.4, xmax=5.7, ymin=2.4, ymax=5.7,
    xtick={3,4,5}, ytick={3,4,5},
    x tick label style={font=\tiny}, y tick label style={font=\tiny},
    grid=major, grid style={gray!15},
    clip=false,
]
\fill[blue!5] (axis cs:4.0,4.5) rectangle (axis cs:5.7,5.7);
\node[font=\tiny, text=blue!55, anchor=north east] at (axis cs:5.65,5.65) {act first};
\addplot[only marks, mark=*, mark size=3.4pt, color=blue!70, fill=blue!45,
         fill opacity=0.75] coordinates {
  (3,5) (5,4) (5,5) (4,4) (4,3) (4.10,4) (3.90,4) (3,3) (3.10,5) (3,4)};
\node[font=\tiny, anchor=east]  at (axis cs:2.88,5.00) {G1};
\node[font=\tiny, anchor=south] at (axis cs:5.00,4.16) {G2};
\node[font=\tiny, anchor=east]  at (axis cs:4.88,5.00) {G3};
\node[font=\tiny, anchor=south] at (axis cs:4.00,4.18) {G4};
\node[font=\tiny, anchor=south] at (axis cs:4.00,3.16) {G5};
\node[font=\tiny, anchor=west]  at (axis cs:4.20,3.86) {G6};
\node[font=\tiny, anchor=east]  at (axis cs:3.80,3.86) {G7};
\node[font=\tiny, anchor=north] at (axis cs:3.00,2.86) {G8};
\node[font=\tiny, anchor=west]  at (axis cs:3.22,5.06) {G9};
\node[font=\tiny, anchor=east]  at (axis cs:2.88,4.00) {G10};
\end{axis}
\end{tikzpicture}
\caption{Each gap of Table~\ref{tab:gaps} plotted by how much it matters against how readily it could be closed. Where two share a cell, one is offset so neither is hidden. Shading marks the quadrant worth attacking first.}
\label{fig:gaps}
\end{figure}
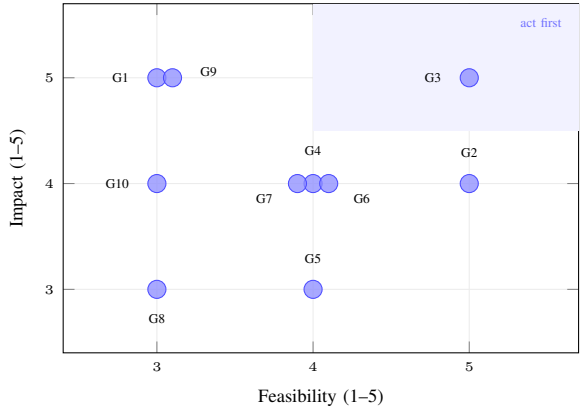

Taken together, the seven paradigms leave ten deficiencies that recur across more than one of them. Table~\ref{tab:gaps} lists these; Figure~\ref{fig:gaps} places each on axes of consequence and tractability. G3 scores highest on both axes: nothing else can be compared until reporting is standardised, and standardising it demands no new technology. Gaps G1 and G9 carry the highest impact but lower feasibility, since closing them requires either economic modelling the field has not attempted or re-running published work on shared environments.

\begin{table}[!t]
\centering
\caption{Ten recurring gaps with consequence and tractability scores.}
\label{tab:gaps}
\small
\renewcommand{\arraystretch}{1.0}
\begin{tabular}{@{}p{0.6cm}p{4.6cm}cc@{}}
\toprule
\textbf{ID} & \textbf{Gap} & \textbf{Imp.} & \textbf{Feas.} \\
\midrule
G1 & Optimizer--load loop left open & 5 & 3 \\
G2 & Validation confined to simulation & 4 & 5 \\
G3 & No standardised reporting or metrics & 5 & 5 \\
G4 & Water excluded from control objectives & 4 & 4 \\
G5 & Embodied carbon excluded & 3 & 4 \\
G6 & Space and time flexibility studied apart & 4 & 4 \\
G7 & AI load dynamics violate load assumptions & 3 & 4 \\
G8 & Safe RL under restrictive thermal models & 3 & 3 \\
G9 & No retrospective re-evaluation on benchmarks & 5 & 3 \\
G10 & Explainability not transferred from grids & 4 & 3 \\
\bottomrule
\end{tabular}
\end{table}

\subsection{Problem Formulation}

\begin{definition}[Net-Benefit Data Center Optimisation]
Given a facility operating under an incumbent controller $\pi_0$ and a workload demand process $\mu_0$, and a candidate learned policy $\pi$ subject to a thermal-safety constraint set $\Phi$, the problem is to report, for an evaluation window: (i) the direct saving achieved at fixed demand; (ii) the induced demand attributable to the efficiency gain, parameterised by an elasticity $\varepsilon$; (iii) the resulting net effect across energy, operational carbon, water withdrawal and the embodied carbon $\Xi$ attributable to the window; and (iv) a record $\sigma$ carrying the validation venue $\nu$ and measurement boundary under which the result was obtained.
\end{definition}

Writing $\delta$ for the relative reduction in the effective unit cost of compute that the efficiency gain produces, and $\mu_\varepsilon$ for the adjusted demand, the net energy benefit is

\begin{equation}
\Delta E_{\mathrm{net}} =
\underbrace{\big[E(\pi_0,\mu_0) - E(\pi,\mu_0)\big]}_{\text{reported today}}
-
\underbrace{\big[E(\pi,\mu_\varepsilon) - E(\pi,\mu_0)\big]}_{\text{rebound term}},
\label{eq:netbenefit}
\end{equation}
\begin{equation}
\mu_\varepsilon = \mu_0\,(1 + \varepsilon\delta).
\label{eq:demand}
\end{equation}

The first bracket is what the literature reports. The second, which the demand adjustment of \eqref{eq:demand} drives, is absent from every control study coded in Section~VI. When $\varepsilon = 0$ the definition collapses to the conventional formulation, which makes the standard result a special case rather than a rival. The same decomposition applies to carbon, water and, with $\Xi$ amortised over the window, to embodied impact; the multi-resource objective is then to report the vector $(\Delta E_{\mathrm{net}}, \Delta C_{\mathrm{net}}, \Delta W_{\mathrm{net}}, \Xi)$ subject to $\Phi$ rather than to optimise any single component.

\textbf{Scope Clarification}: This is a specification for how results should be reported and evaluated, not a solved optimisation with convergence guarantees. In particular, $\varepsilon$ is not identified by any dataset in this corpus; the definition makes its absence visible rather than supplying its value.

\subsection{CLEAR-DC Architecture}

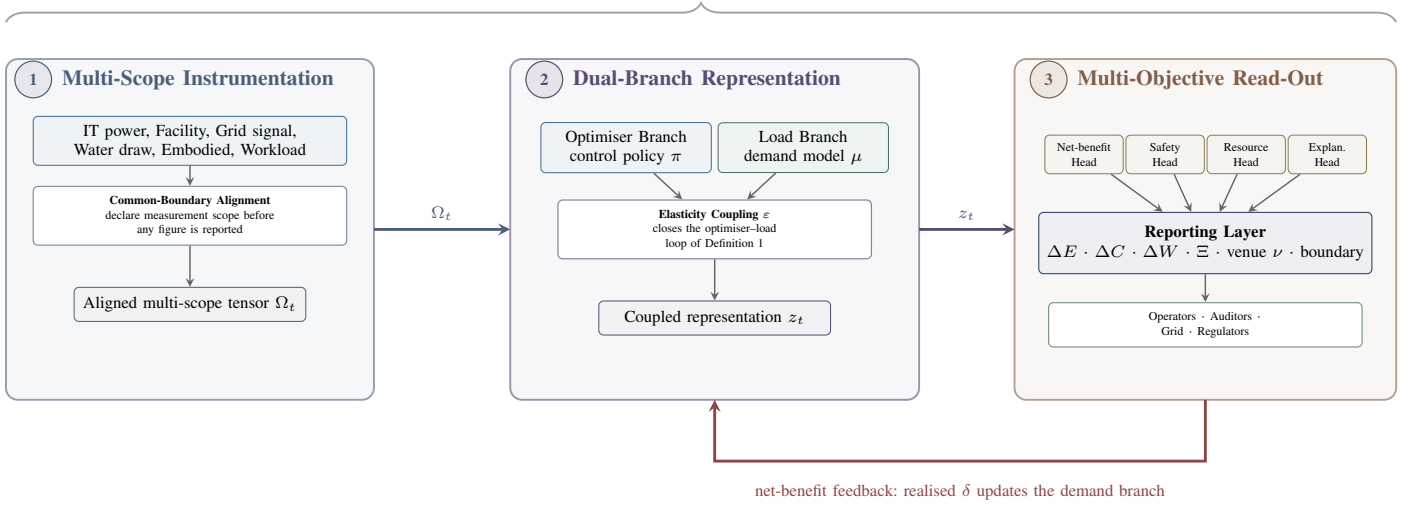
\begin{figure*}[!t]
\centering
\resizebox{\textwidth}{!}{%
\begin{tikzpicture}
\begin{scope}[shift={(0,0)}]
    \node[stagebox=stage1Color, minimum width=5.4cm, minimum height=5.0cm] (s1bg) at (2.4,0) {};
    \node[stagenum=stage1Color] at (0.1,2.2) {1};
    \node[font=\small\bfseries, text=stage1Color, right] at (0.4,2.2) {Multi-Scope Instrumentation};
    \node[procbox=dataColor, minimum width=4.6cm] (inputs) at (2.4,1.3) {IT power, Facility, Grid signal,\\Water draw, Embodied, Workload};
    \node[eqbox=stage1Color, minimum width=4.6cm] (align) at (2.4,0.2) {\textbf{Common-Boundary Alignment}\\declare measurement scope before\\any figure is reported};
    \node[procbox=stage1Color, minimum width=3.4cm] (out1) at (2.4,-1.1) {Aligned multi-scope tensor $\Omega_t$};
    \draw[flowarrow] (inputs) -- (align);
    \draw[flowarrow] (align) -- (out1);
\end{scope}
\begin{scope}[shift={(7.5,0)}]
    \node[stagebox=stage2Color, minimum width=6.0cm, minimum height=5.0cm] (s2bg) at (2.6,0) {};
    \node[stagenum=stage2Color] at (0.1,2.2) {2};
    \node[font=\small\bfseries, text=stage2Color, right] at (0.4,2.2) {Dual-Branch Representation};
    \node[procbox=dataColor, minimum width=2.5cm] (optb) at (1.3,1.2) {Optimiser Branch\\control policy $\pi$};
    \node[procbox=policyColor, minimum width=2.5cm] (loadb) at (3.9,1.2) {Load Branch\\demand model $\mu$};
    \node[eqbox=fusionColor, minimum width=4.6cm] (elas) at (2.6,0.0) {\textbf{Elasticity Coupling} $\varepsilon$\\closes the optimiser--load\\loop of Definition~1};
    \node[procbox=fusionColor, minimum width=3.4cm] (out2) at (2.6,-1.3) {Coupled representation $z_t$};
    \draw[flowarrow] (optb) -- (elas);
    \draw[flowarrow] (loadb) -- (elas);
    \draw[flowarrow] (elas) -- (out2);
\end{scope}
\begin{scope}[shift={(14.9,0)}]
    \node[stagebox=stage3Color, minimum width=5.6cm, minimum height=5.0cm] (s3bg) at (2.4,0) {};
    \node[stagenum=stage3Color] at (0.1,2.2) {3};
    \node[font=\small\bfseries, text=stage3Color, right] at (0.4,2.2) {Multi-Objective Read-Out};
    \node[procbox=outputColor, minimum width=1.15cm, font=\tiny] (nb)  at (0.62,1.1) {Net-benefit\\Head};
    \node[procbox=outputColor, minimum width=1.15cm, font=\tiny] (saf) at (1.81,1.1) {Safety\\Head};
    \node[procbox=outputColor, minimum width=1.15cm, font=\tiny] (res) at (3.00,1.1) {Resource\\Head};
    \node[procbox=outputColor, minimum width=1.15cm, font=\tiny] (xai) at (4.19,1.1) {Explan.\\Head};
    \node[procbox=navyColor, minimum width=4.6cm, minimum height=0.9cm] (rep) at (2.4,-0.2) {\textbf{Reporting Layer}\\$\Delta E$ $\cdot$ $\Delta C$ $\cdot$ $\Delta W$ $\cdot$ $\Xi$ $\cdot$ venue $\nu$ $\cdot$ boundary};
    \node[eqbox=successColor, minimum width=4.6cm] (users) at (2.4,-1.4) {Operators $\cdot$ Auditors $\cdot$\\Grid $\cdot$ Regulators};
    \draw[flowarrow] (nb)  -- (rep);
    \draw[flowarrow] (saf) -- (rep);
    \draw[flowarrow] (res) -- (rep);
    \draw[flowarrow] (xai) -- (rep);
    \draw[flowarrow] (rep) -- (users);
\end{scope}
\draw[stagearrow, color=stage1Color] (5.1,0) -- (7.1,0) node[midway, above, font=\scriptsize] {$\Omega_t$};
\draw[stagearrow, color=stage2Color] (13.1,0) -- (14.5,0) node[midway, above, font=\scriptsize] {$z_t$};
\draw[stagearrow, color=gapColor] (17.3,-2.5) -- (17.3,-3.4) -- (10.1,-3.4) -- (10.1,-2.5);
\node[font=\scriptsize, text=gapColor, anchor=north] at (13.7,-3.62)
    {net-benefit feedback: realised $\delta$ updates the demand branch};
\path (-0.3,-4.1) rectangle (20.1,-4.1);
\draw[decorate, decoration={brace, amplitude=8pt, raise=4pt}, line width=0.8pt, contextColor!60]
    (-0.3, 2.9) -- (20.1, 2.9)
    node[midway, above=0.45cm, font=\small\bfseries, text=black] {CLEAR-DC: Closed-Loop Energy Accounting and Reporting for Data Centers};
\end{tikzpicture}}
\caption{CLEAR-DC: six instrumentation scopes are aligned onto a declared boundary, encoded by an optimiser branch and a load branch coupled through an explicit elasticity term, and read out by four heads into a schema-conformant record.}
\label{fig:overview}
\end{figure*}

Figure~\ref{fig:overview} lays out CLEAR-DC (Closed-Loop Energy Accounting and Reporting for Data Centers) in three stages. Stage~1 instruments the facility across six scopes --- information technology power, facility overhead, grid signals, water withdrawal, embodied hardware impact and workload composition --- and aligns them onto a declared common measurement boundary, so that a reported figure carries the scope it was measured over. Stage~2 encodes the control problem and the demand problem in separate branches: an optimiser branch holding the policy $\pi$, realisable by any sequence model that exposes attention weights or a comparable attribution signal, and a load branch representing the demand process $\mu$. The two are joined by an elasticity coupling that carries the efficiency gain from the optimiser branch into the demand branch, which is the component that closes the loop of Definition~1.

Stage~3 reads out four heads from the coupled representation. The net-benefit head evaluates \eqref{eq:netbenefit} rather than the first bracket alone. The safety head enforces $\Phi$ by projection, following the rectification approach that has proved effective in constrained cooling control \cite{ref55,ref4}. The resource head reports water and embodied components alongside energy. The explanation head produces attributions with a stability check across resampled backgrounds, since an unstable explanation is not auditable. All four feed a reporting layer that emits the record $\sigma$ in the schema of Section~IV.

Algorithm~\ref{alg:cleardc} specifies the proposed evaluation for one control interval.

\begin{algorithm}[t]
\caption{CLEAR-DC net-benefit evaluation, one control interval.}
\label{alg:cleardc}
\small
\textbf{Notation}: $\Omega_t$: instrumented window; $B$: resamples for attribution stability.
\begin{algorithmic}[1]
\Require Window $\Omega_t$, incumbent $\pi_0$, policy $\pi$, constraints $\Phi$, elasticity prior $\varepsilon$
\Ensure Net-benefit vector, safe action $\hat{a}$, record $\sigma$
\State $\bar{\Omega} \gets \text{AlignBoundary}(\Omega_t)$ \Comment{Stage 1}
\State $z^{P} \gets \text{OptimiserBranch}(\bar{\Omega}; \theta_P)$ \Comment{Stage 2}
\State $z^{L} \gets \text{LoadBranch}(\bar{\Omega}; \theta_L)$ \Comment{Stage 2}
\State $\delta \gets \text{UnitCostDelta}(z^{P}, \pi_0)$
\State $z_t \gets \text{Couple}(z^{P}, z^{L}, \varepsilon, \delta)$ \Comment{closes the loop}
\State $a \gets \text{PolicyHead}(z_t)$
\State $\hat{a} \gets \text{Project}(a, \Phi)$ \Comment{safety head}
\State $(\Delta E, \Delta C, \Delta W) \gets \text{NetBenefit}(z_t, \varepsilon, \delta)$ \Comment{Eq.~\eqref{eq:netbenefit}}
\State $\Xi \gets \text{AmortiseEmbodied}(\bar{\Omega})$ \Comment{resource head}
\For{$b = 1, \ldots, B$}
    \State $A_b \gets \text{Attribute}(z_t, \text{resample}_b)$
\EndFor
\State $A \gets \text{StabilityCheck}(\{A_b\}_{b=1}^{B})$ \Comment{explanation head}
\State $\sigma \gets \text{Record}(\Delta E, \Delta C, \Delta W, \Xi, A, \nu, \text{boundary})$
\State \Return $(\hat{a}, \sigma)$ to the reporting layer
\end{algorithmic}
\end{algorithm}

\subsection{From Gaps to Components}

Every component of Figure~\ref{fig:overview} exists because of a specific entry in Table~\ref{tab:gaps}. Table~\ref{tab:traceability} records which, so a reader can check the design against the diagnosis instead of taking it on trust.

\begin{table}[!t]
\centering
\caption{Each component and the gaps it responds to.}
\label{tab:traceability}
\small
\renewcommand{\arraystretch}{1.05}
\begin{tabular}{@{}p{2.1cm}p{1.05cm}p{4.0cm}@{}}
\toprule
\textbf{Component} & \textbf{Gaps} & \textbf{Rationale} \\
\midrule
Boundary alignment & G3 & Forces the measurement scope to be declared before any figure is reported \\
Water, embodied scopes & G4, G5 & Instruments resources the coded corpus never carries into control \\
Optimiser branch & G8, G10 & Exposes an attribution signal; safety set applied to its output \\
Load branch & G1, G7 & Represents demand as a modelled process, including AI load dynamics \\
Elasticity coupling & G1 & The rebound term of Eq.~\eqref{eq:netbenefit}, absent from prior formulations \\
Net-benefit head & G1, G3 & Reports the decomposition, not the direct bracket alone \\
Safety head & G8 & Projection onto $\Phi$ rather than reward shaping \\
Resource head & G4, G5, G6 & Vector read-out enables joint space--time--resource objectives \\
Explanation head & G10 & Attribution with a stability check across resamples \\
Reporting layer & G2, G3, G9 & Emits venue and boundary, enabling retrospective comparison \\
\bottomrule
\end{tabular}
\end{table}

\section{How the Framework Should Be Evaluated}

\subsection{A Design Case: India's Data Center Build-Out}

India is among the fastest-growing data center markets, with concentrated capacity in Mumbai, Chennai and Hyderabad and a grid whose carbon intensity varies substantially by region and hour. Several conditions make it a useful design case rather than a benchmark. Water stress is spatially uneven and, in several hosting regions, severe, which activates the tradeoff documented between electrical efficiency and water withdrawal in hot-arid climates \cite{ref19,ref23}. Grid carbon intensity varies enough to make temporal shifting meaningful, while the evidence that flexibility can raise emissions under low renewable penetration \cite{ref49} means the direction of benefit cannot be assumed. Table~\ref{tab:datasources} lists candidate data categories for an instantiation. We stress that this is a design case: no results are reported for it, and public operational data at the granularity CLEAR-DC requires is not currently available.

\begin{table}[!t]
\centering
\caption{Data categories an India deployment would require.}
\label{tab:datasources}
\small
\renewcommand{\arraystretch}{1.1}
\begin{tabular}{@{}p{1.8cm}p{5.4cm}@{}}
\toprule
\textbf{Category} & \textbf{Candidate sources} \\
\midrule
Facility & Operator PUE and WUE disclosures; colocation reports \\
Grid & Central Electricity Authority generation mix and CO$_2$ baseline database; regional load dispatch data \\
Climate & India Meteorological Department station records; wet-bulb series for economiser feasibility \\
Water & Central Ground Water Board stress classification; municipal supply tariffs \\
Workload & Public cluster traces as proxy; accelerator power profiles from measurement studies \cite{ref22} \\
Embodied & Hardware lifecycle inventories; refresh cycle assumptions \\
Policy & MeitY data center policy instruments; state incentive frameworks \\
\bottomrule
\end{tabular}
\end{table}

\subsection{A Minimum Set of Reported Fields}

Table~\ref{tab:metrics} sets out the fields we argue should be reported together, beside what is reported at present. The schema is deliberately modest: every field is either already measured by some subset of the corpus or derivable from information the authors necessarily possess. Its value is not novelty but joint reporting, since it is the absence of the venue and boundary fields that makes the savings intervals of Section~VI uninterpretable.

\begin{table}[!t]
\centering
\caption{Reporting fields proposed here, with their present status.}
\label{tab:metrics}
\small
\renewcommand{\arraystretch}{1.1}
\begin{tabular}{@{}p{2.1cm}p{2.6cm}p{2.9cm}@{}}
\toprule
\textbf{Field} & \textbf{Content} & \textbf{Current status} \\
\midrule
Baseline & Identity of $\pi_0$ (rule-based, PID, MPC, prior policy) & Reported, but often unspecified \\
Boundary & IT-only, cooling subsystem, facility, or lifecycle & Rarely stated explicitly \\
Venue $\nu$ & Simulation, testbed, field, production & Inferable at best; never a field \\
Direct saving & $\Delta E$ at fixed demand & Universally reported \\
Net saving & $\Delta E_{\mathrm{net}}$ after induced demand & Not reported anywhere in this corpus \\
Water & $\Delta W$ and basin stress class & Only in macro studies \\
Embodied & $\Xi$ amortised over window & Only in AI-load studies \\
Safety & Violations of $\Phi$ during learning and operation & Reported by safe-RL work only \\
Explanation & Attribution stability across resamples & Essentially untested \\
\bottomrule
\end{tabular}
\end{table}

\subsection{Scoring the Paradigms}

\begin{figure}[!t]
\centering
\begin{tikzpicture}
\begin{axis}[
    ybar, bar width=2.1pt,
    width=8.6cm, height=5.4cm,
    ylabel={Qualitative score (1--5)},
    ylabel style={font=\scriptsize},
    symbolic x coords={Saving,Deploy,Safety,Breadth,Explain,Loop},
    xtick=data,
    xticklabels={Saving, Deploy, Safety, Breadth, Explain, Loop},
    x tick label style={font=\tiny, rotate=25, anchor=north east},
    y tick label style={font=\tiny},
    ymin=0, ymax=5.6,
    legend style={font=\tiny, at={(0.5,-0.30)}, anchor=north, legend columns=2, draw=none},
    enlarge x limits=0.12,
]
\addplot[fill=gray!55]   coordinates {(Saving,2) (Deploy,3) (Safety,1) (Breadth,1) (Explain,3) (Loop,1)};
\addplot[fill=blue!55]   coordinates {(Saving,4) (Deploy,2) (Safety,1) (Breadth,1) (Explain,1) (Loop,1)};
\addplot[fill=teal!65]   coordinates {(Saving,4) (Deploy,2) (Safety,4) (Breadth,1) (Explain,2) (Loop,1)};
\addplot[fill=orange!65] coordinates {(Saving,3) (Deploy,4) (Safety,2) (Breadth,2) (Explain,2) (Loop,1)};
\addplot[fill=red!55]    coordinates {(Saving,4) (Deploy,4) (Safety,4) (Breadth,5) (Explain,4) (Loop,5)};
\legend{Predictive ML, Model-free DRL, Safe/constrained RL, Carbon-aware, Proposed (target)}
\end{axis}
\end{tikzpicture}
\caption{Six axes, five paradigms. Bars for the proposed framework record intentions, not measurements.}
\label{fig:paradigms}
\end{figure}
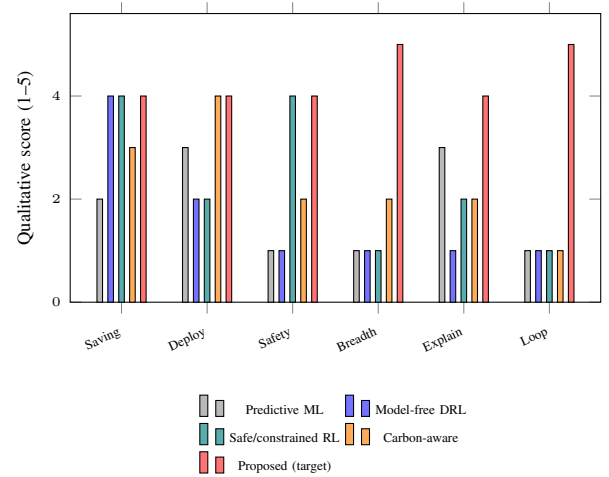

Figure~\ref{fig:paradigms} scores the paradigms of Table~\ref{tab:related_work_comparison} on the six axes implied by Table~\ref{tab:metrics}, with CLEAR-DC's intended position shown alongside. \textbf{These scores are our reading of the literature and rest on no experiment}; the bars for the proposed framework state what it is meant to achieve, and quoting them as achieved performance would misrepresent this paper.

\section{Reproducibility and Artefact Availability}

The empirical claims in this paper are claims about a corpus, so the corpus is the artefact. The coding sheet, the derived statistics and the code that computes every count in Section~VI are released at:

\begingroup\sloppy\footnotesize\noindent
\url{https://github.com/Kimalice/AI-for-Energy-Optimization-in-Data-Centers-Closing-the-Optimizer-Load-Loop}
\par\endgroup
\vspace{2pt}

We report the following:

\begin{itemize}[leftmargin=*]
    \item \textbf{Search record}: All twenty queries are listed with their filters, returned counts and status. Retrieval was performed on a single date; the access tier used caps results at ten per query, which bounds coverage at roughly 200 records before deduplication.
    \item \textbf{Coding sheet}: Each of the 63 retained papers carries a venue label, a scope label and two binary resource flags. The full sheet, including the coding rules and the ambiguous cases, accompanies this paper.
    \item \textbf{Derived statistics}: Every count in Section~VI is computed from the coding sheet by a short script rather than tallied by hand, so the numbers can be regenerated and disagreements localised to individual coding decisions.
    \item \textbf{Figure provenance}: Figures~\ref{fig:venue} and~\ref{fig:ranges} are generated directly from the coding sheet and from the intervals reported in \cite{ref39} respectively; no value in either is estimated.
\end{itemize}

\section{Corpus Analysis: First Quantitative Findings}

The 63 coded papers comprise 13 secondary studies and 50 primary studies. Of the primary studies, 28 are control-oriented, meaning their principal contribution is a method that acts on a facility through cooling control, scheduling, or carbon-aware placement. Table~\ref{tab:corpus} collects the counts; the findings below concern that subset unless stated otherwise, and are reported as coded.

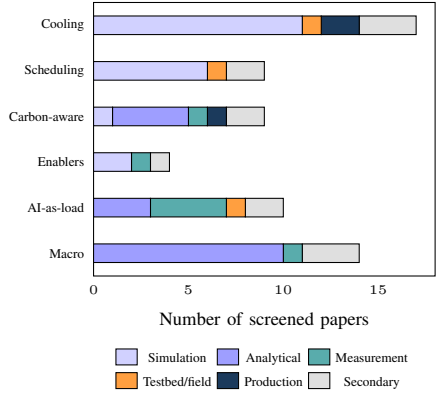
\begin{figure}[!t]
\centering
\begin{tikzpicture}
\begin{axis}[
    xbar stacked, bar width=7pt,
    width=6.1cm, height=5.2cm,
    xlabel={Number of screened papers},
    xlabel style={font=\scriptsize},
    symbolic y coords={Macro,AI-as-load,Enablers,Carbon-aware,Scheduling,Cooling},
    ytick=data,
    y tick label style={font=\tiny},
    x tick label style={font=\tiny},
    xmin=0, xmax=18,
    legend style={font=\tiny, at={(0.5,-0.24)}, anchor=north, legend columns=3, draw=none},
    tick style={draw=none},
]
\addplot[fill=blue!18] coordinates {(11,Cooling) (6,Scheduling) (1,Carbon-aware) (2,Enablers) (0,AI-as-load) (0,Macro)};
\addplot[fill=blue!38] coordinates {(0,Cooling) (0,Scheduling) (4,Carbon-aware) (0,Enablers) (3,AI-as-load) (10,Macro)};
\addplot[fill=teal!65] coordinates {(0,Cooling) (0,Scheduling) (1,Carbon-aware) (1,Enablers) (4,AI-as-load) (1,Macro)};
\addplot[fill=orange!75] coordinates {(1,Cooling) (1,Scheduling) (0,Carbon-aware) (0,Enablers) (1,AI-as-load) (0,Macro)};
\addplot[fill=navyColor] coordinates {(2,Cooling) (0,Scheduling) (1,Carbon-aware) (0,Enablers) (0,AI-as-load) (0,Macro)};
\addplot[fill=gray!25] coordinates {(3,Cooling) (2,Scheduling) (2,Carbon-aware) (1,Enablers) (2,AI-as-load) (3,Macro)};
\legend{Simulation, Analytical, Measurement, Testbed/field, Production, Secondary}
\end{axis}
\end{tikzpicture}
\caption{Validation venue by scope across the 63 coded papers. Production evidence is confined almost entirely to cooling, and the AI-as-load strand is dominated by measurement rather than control.}
\label{fig:venue}
\end{figure}

\textit{Validation venue.} Eighteen of the 28 control-oriented studies, or 64 percent, are validated in simulation or trace-driven replay alone. Two use a physical testbed or field deployment and three report results from a production facility, giving five studies, or 18 percent, with evidence from real infrastructure. The remaining five are analytical or measurement studies without a controller in the loop. Figure~\ref{fig:venue} shows the distribution by scope and makes the asymmetry visible: production evidence exists almost entirely in the cooling paradigm, and is concentrated in work originating from operators rather than universities \cite{ref13,ref60,ref41}. This is consistent with, and quantifies for data centers specifically, the deployment shortfall reported for building control more broadly \cite{ref2}. Twenty-eight is a small denominator, and the released analysis prices that rather than leaving it to the reader: resampling the coded corpus places a 95 percent interval of 46 to 82 percent around the simulation-only share, and reassigning two venue codes would carry the point estimate below 60 percent. The direction of the result is secure at this sample size; the second significant figure is not.

\textit{Resource breadth.} Nine of the 63 papers account for water withdrawal and six account for embodied carbon. None of the 28 control-oriented studies does either. Water appears exclusively in macro-estimation and measurement work \cite{ref34,ref51,ref23,ref19,ref17}, and embodied carbon exclusively in the AI-as-load strand \cite{ref28,ref9,ref58,ref32}. The separation is clean enough to be structural rather than incidental: the studies that measure these resources do not build controllers, and the studies that build controllers do not measure these resources. A policy minimising facility energy therefore optimises against an objective that provably excludes two of the four impact channels its own motivating literature identifies.

\begin{figure}[!t]
\centering
\begin{tikzpicture}
\begin{axis}[
    xbar stacked, bar width=8pt,
    width=6.0cm, height=4.4cm,
    xlabel={Reported energy saving vs.\ baseline (\%)},
    xlabel style={font=\scriptsize},
    symbolic y coords={DRL,Meta,Heur,ML,Stat},
    ytick=data,
    yticklabels={DRL cooling (corpus), Metaheuristic, Heuristic, Machine learning, Statistical},
    y tick label style={font=\tiny},
    x tick label style={font=\tiny},
    xmin=0, xmax=100,
    tick style={draw=none},
]
\addplot[draw=none, fill=none] coordinates {(5.4,Stat) (1.6,ML) (5.4,Heur) (7.68,Meta) (7.1,DRL)};
\addplot[fill=blue!45] coordinates {(78.6,Stat) (86.9,ML) (84.6,Heur) (89.32,Meta) (0,DRL)};
\addplot[fill=navyColor] coordinates {(0,Stat) (0,ML) (0,Heur) (0,Meta) (19.5,DRL)};
\end{axis}
\end{tikzpicture}
\caption{Reported energy savings intervals. The four upper bars are the ranges recorded across technique families in a systematic review; the lower bar is the span of DRL cooling results in this corpus. Overlap is near-total.}
\label{fig:ranges}
\end{figure}
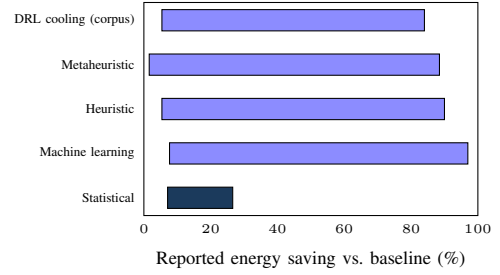

\textit{The savings-range problem.} Figure~\ref{fig:ranges} plots the intervals reported by a systematic review across four technique families \cite{ref39} against the range spanned by the deep reinforcement learning cooling studies in this corpus. Three observations follow. The four review intervals overlap across almost their entire extent, so no ordering of technique families is supported. The intervals are also far wider than any plausible physical range for a single facility, which indicates that they aggregate across incommensurable baselines and boundaries rather than measuring method quality. And the corpus range for cooling control, roughly 7 to 27 percent, is both narrower and lower than the upper reaches of the review intervals --- not because those methods are better, but because the studies behind the wider intervals report against weaker baselines over narrower boundaries.

\textit{Temporal structure.} Ten papers form the AI-as-load strand, and nine of the ten were published in 2023 or later. That strand is therefore almost entirely younger than the control literature it needs to inform, which offers a benign explanation for the disjunction reported above --- the two bodies of work have barely had time to meet. It also implies that the disjunction is tractable, since no entrenched methodological commitment separates them.

\begin{table}[!t]
\centering
\caption{Coded corpus, summary counts.}
\label{tab:corpus}
\small
\renewcommand{\arraystretch}{1.05}
\begin{tabular}{@{}p{4.6cm}cc@{}}
\toprule
\textbf{Dimension} & \textbf{Count} & \textbf{Share} \\
\midrule
Papers coded & 63 & --- \\
Secondary studies & 13 & 21\% of 63 \\
Control-oriented primary studies & 28 & 44\% of 63 \\
\quad simulation or trace only & 18 & 64\% of 28 \\
\quad testbed or field deployment & 2 & 7\% of 28 \\
\quad production facility & 3 & 11\% of 28 \\
\quad accounting for water & 0 & 0\% of 28 \\
\quad accounting for embodied carbon & 0 & 0\% of 28 \\
\quad reporting net of induced demand & 0 & 0\% of 28 \\
AI-as-load studies & 10 & 16\% of 63 \\
\quad published 2023 or later & 9 & 90\% of 10 \\
Papers accounting for water & 9 & 14\% of 63 \\
Papers accounting for embodied carbon & 6 & 10\% of 63 \\
\bottomrule
\end{tabular}
\end{table}

\textit{Reading these results.} These counts do not show that learned control does not work; several of the coded studies report credible savings under credible constraints. They show that the field's evidence base has a shape, and that the shape is narrow in three specific ways --- venue, resource scope, and demand framing --- each of which corresponds to a gap in Table~\ref{tab:gaps}. The purpose of the schema in Table~\ref{tab:metrics} is to make that shape visible in future work at the cost of a few additional reported fields.

\section{Discussion}

\subsection{What This Paper Establishes}

No method in Table~\ref{tab:related_work_comparison} holds all five properties at once: joint optimisation, safety guarantees, resource breadth, loop closure, and evidence from outside a simulator. CLEAR-DC is drawn to fill that empty cell. We do not claim it succeeds in doing so; settling that requires building it. What this paper does establish empirically is narrower and, we would argue, more useful: the distribution of the evidence base reported in Section~VI, and the observation that the two literatures most relevant to the question are structurally disjoint.

\subsection{Before Any Implementation}

Several design choices are worth surfacing before any implementation. We treat the safety head as mandatory rather than optional, following evidence that reward shaping learns constraint compliance only after experiencing violations \cite{ref55}. The explanation head is included because operator trust, not algorithmic performance, appears to be the binding constraint on adoption, and because the explanation techniques most likely to be reached for are known to be locally unstable unless checked. The elasticity coupling introduces a parameter that cannot currently be estimated from any dataset in this corpus; in an implementation it would be supplied as a scenario range rather than a point value, and results reported across that range.

\subsection{Limits of the Evidence}

\begin{enumerate}[leftmargin=*]
    \item \textit{What the evidence supports}: the quantitative component counts publications; it does not measure facilities. It characterises what has been published, not what is physically true of data centers. A framework evaluated as unpublished or proprietary work would be invisible to it, and industrial results are systematically under-published.
    \item \textit{Retrieval bound}: twenty queries at ten results each bound coverage at roughly 200 records. Sub-areas addressed by a single query are correspondingly thin, and one query intended to cover edge deployment returned largely mobile federated-learning work, so edge scenarios are under-represented.
    \item \textit{Coding method}: coding was performed at abstract level by a single coder. A paper that accounts for water only in its full text would be coded as not accounting for it. Double coding with an inter-rater statistic would strengthen the resource-breadth claims in particular.
    \item \textit{Prioritization method}: the impact and feasibility scores in Table~\ref{tab:gaps} reflect the authors' synthesis rather than formal elicitation; a structured expert panel would place them on firmer ground.
    \item \textit{Unidentified elasticity}: Definition~1 introduces $\varepsilon$ without supplying it. The definition makes the omission explicit and bounded rather than resolving it, and a net-benefit figure computed with an assumed $\varepsilon$ should be read as a sensitivity, not a measurement.
    \item \textit{Framework not implemented}: CLEAR-DC has not been built, trained or benchmarked. The bars in Figure~\ref{fig:paradigms} are design targets, and no performance claim is made for it.
\end{enumerate}

\subsection{Broader Impact and Ethical Considerations}

Three considerations warrant emphasis. \textit{Resource justice}: a controller optimising electricity alone may increase water withdrawal in a stressed basin, and the populations affected by that withdrawal are not parties to the optimisation; the resource head exists for this reason. \textit{Measurement honesty}: an efficiency claim that omits induced demand can support a decision to expand capacity that the underlying analysis does not justify, and the risk grows as such claims enter policy discussion. \textit{Regional asymmetry}: regulators and smaller operators in emerging markets are less equipped to audit a learned controller independently than are the hyperscale operators that produce most of the production evidence in this corpus, which is part of why explanation stability and declared measurement boundaries are treated here as requirements rather than refinements.

\section{Conclusion and Future Work}

Across predictive modelling, reinforcement learning, scheduling and carbon-aware placement, the reported gains rest on a premise this paper has tried to make visible: the field evaluates its controllers against a demand process it holds fixed, over a resource scope that excludes water and embodied impact, in settings that are overwhelmingly simulated. Coding 63 papers makes the shape of that evidence base measurable rather than anecdotal.

\subsection{Summary of Contributions}
\begin{itemize}[leftmargin=*]
    \item A reading of seven paradigms, each closed by naming what it leaves undone (Section~II).
    \item A coded corpus of 63 papers showing that 18 of 28 control studies are simulation-only and none account for water or embodied carbon (Section~VI).
    \item Ten recurring gaps, scored on consequence and tractability and ranked (Section~III).
    \item CLEAR-DC, a framework traceable component by component to those gaps (Section~III).
    \item A minimum reporting schema whose joint adoption would make future results comparable (Section~IV).
\end{itemize}

\subsection{Where to Go Next}
\begin{itemize}[leftmargin=*]
    \item \textbf{Elasticity estimation}: identify $\varepsilon$ empirically from operator data linking efficiency improvements to subsequent capacity provisioning, which would convert Definition~1 from a specification into a measurement.
    \item \textbf{Retrospective benchmarking}: re-evaluate headline results from the coded corpus on the shared environments that appeared in 2024 and 2025 \cite{ref36,ref37}, which is the cheapest available test of whether reported savings survive a common setting.
    \item \textbf{Water-aware control}: extend a safe cooling controller with a water objective and evaluate it across climate zones where the electricity--water tradeoff reverses \cite{ref19}.
    \item \textbf{Load-dynamics-aware control}: relax the smooth-load assumption using measured accelerator power profiles \cite{ref22,ref7} and test whether existing controllers remain stable under realistic training transients.
    \item \textbf{Double-coded replication}: repeat the corpus analysis with two independent coders and full-text access to establish inter-rater reliability for the resource-breadth findings.
\end{itemize}

\subsection{Adopting Parts of the Framework}

CLEAR-DC is specified as an accounting and reporting structure rather than a fixed pipeline. The optimiser branch accepts any policy class exposing an attribution signal, so an existing safe controller can be dropped in unchanged. The load branch accepts any demand model, including a constant one, in which case the framework degrades gracefully to the conventional open-loop formulation. The reporting layer is independent of both and could be adopted on its own by studies that decline the rest of the architecture --- which, given the findings of Section~VI, would be the single most valuable outcome of this paper.

\bibliographystyle{IEEEtran}

\end{document}